\documentclass[10pt,twocolumn,letterpaper]{article}

\usepackage{cvpr}              
\definecolor{cvprblue}{rgb}{0.21,0.49,0.74}
\usepackage[pagebackref,breaklinks,colorlinks,allcolors=cvprblue]{hyperref}
\usepackage{booktabs}
\usepackage{tabularx}
\usepackage{threeparttable}
\usepackage{multirow}
\usepackage{graphicx}
\usepackage{subcaption}
\usepackage{tcolorbox}
\tcbuselibrary{skins, breakable}
\usepackage{algorithm}
\usepackage{algorithmic}

\usepackage{amsmath}
\usepackage{amssymb}
\usepackage{diagbox} 
\usepackage{caption}
\def\confName{CVPR}
\def\confYear{2026}

\title{Morphology-Consistent Humanoid Interaction through Robot-Centric Video Synthesis}

\author{
    Weisheng Xu$^{1 *}$, \quad 
    Jian Li$^{1 *}$, \quad 
    Yi Gu$^1$, \quad 
    Bin Yang$^1$, \quad 
    Haodong Chen$^2$, \\
    Shuyi Lin$^3$, \quad 
    Mingqian Zhou$^4$, \quad 
    Jing Tan$^1$, \quad 
    Qiwei Wu$^1$, \quad 
    Xiangrui Jiang$^1$, \\
    Taowen Wang$^1$, \quad 
    Jiawen Wen$^1$, \quad 
    Qiwei Liang$^1$, \quad 
    Jiaxi Zhang$^1$, \quad 
    Renjing Xu$^{1 \dagger}$ \\
    \and
    $^1$Hong Kong University of Science and Technology (Guangzhou) \\
    $^2$Harbin Institute of Technology, Shenzhen \quad
    $^3$Shenzhen University \quad
    $^4$University of Cambridge \\
{\tt\small wxu421@connect.hkust-gz.edu.cn, renjingxu@hkust-gz.edu.cn} \\
    \vspace{2mm} 
    Project Page: \url{https://wesleyxu224.github.io/Dream2Act/}
}

\begin{document}
\maketitle

\def\thefootnote{*}\footnotetext{Equal contribution.}\def\thefootnote{\arabic{footnote}}
\def\thefootnote{$\dagger$}\footnotetext{Corresponding author.}\def\thefootnote{\arabic{footnote}}
\begin{abstract}
Equipping humanoid robots with versatile interaction skills typically requires either extensive task-specific policy training or explicit human-to-robot motion retargeting.
However, learning-based policies are hindered by prohibitive data collection costs, limiting their scalability. Meanwhile, retargeting paradigms rely heavily on human-centric pose estimation (e.g., SMPL), which inevitably introduces the morphological gap. Such skeletal scale mismatches result in severe spatial misalignments when mapped to robots, compromising interaction success. 
In this work, we propose Dream2Act, a robot-centric framework that enables zero-shot interaction through generative video synthesis. 
Given an image of the target robot and target object in third-person view, our framework leverages video generation models to synthesize video sequences in which the physical robot completes the task with spatially aligned, morphology-consistent motion. 
We then employ a robust, high-fidelity pose extraction system to recover physically feasible, robot-native joint trajectories from these synthesized ``dreams'', which are subsequently executed via a general-purpose whole-body controller. By operating strictly within the robot-native coordinate space, Dream2Act effectively avoids traditional retargeting errors and eliminates the need for task-specific policy training. 
We evaluate Dream2Act on the Unitree G1 across four categories of whole-body mobile interaction tasks (multi-position ball kicking, sofa sitting, bag punching, and box hugging). Dream2Act achieves an overall task success rate of $37.5\%$, compared to $0\%$ for conventional retargeting pipelines. Retargeting frequently fails to establish correct physical contacts due to the morphological gap, with errors further compounded under locomotion. In contrast, Dream2Act maintains robot-consistent spatial alignment throughout execution, enabling reliable contact formation and substantially higher task completion.
\end{abstract}    
\section{Introduction}
\label{sec:intro}
\begin{figure*}[h!]
    \centering
    \includegraphics[width=\textwidth]{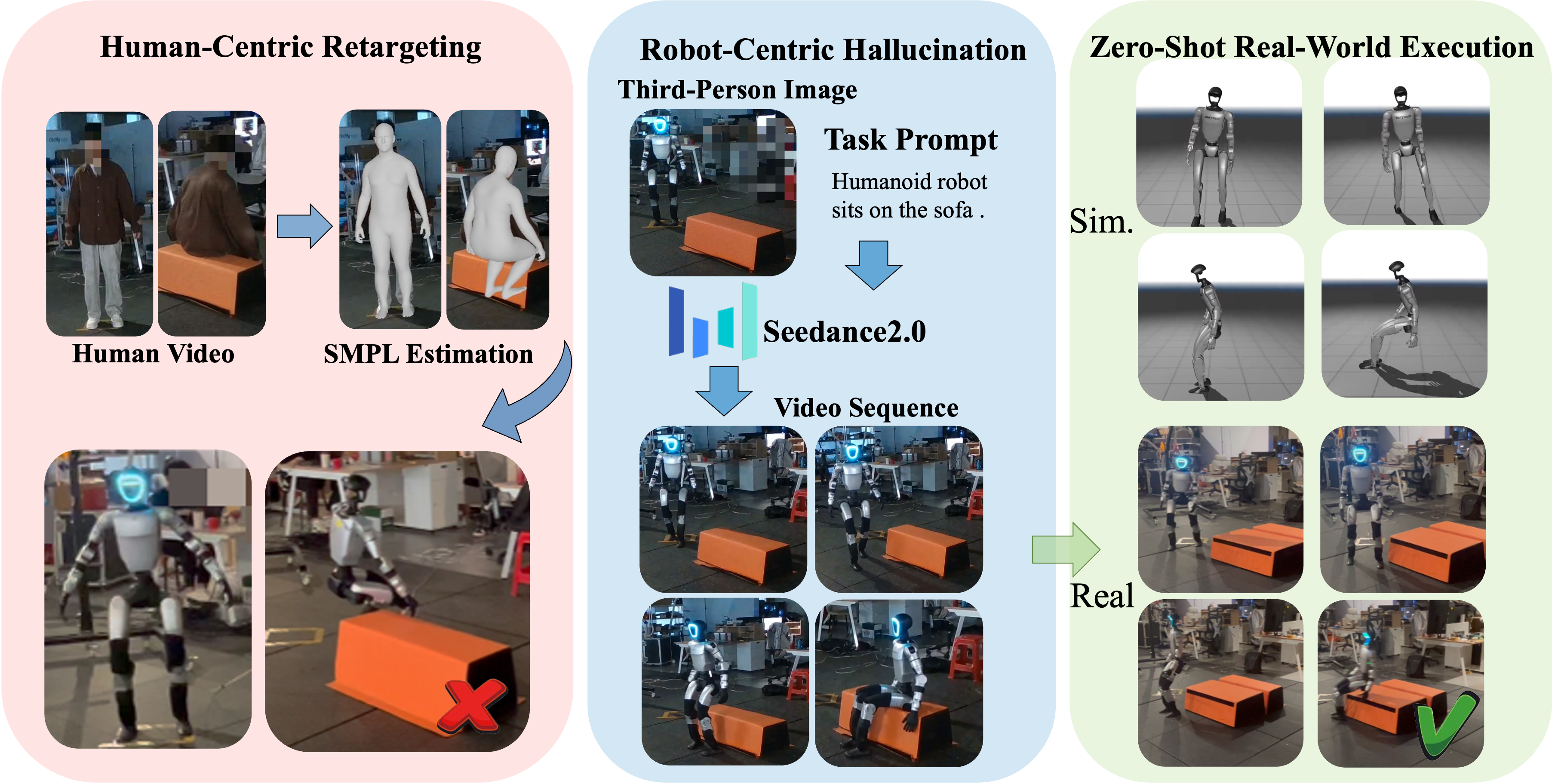}
    \caption{\textbf{Dream2Act: Zero-shot Humanoid Interaction via Robot-Centric Video Synthesis.} (A) Traditional human-centric retargeting fails due to the morphology gap, causing spatial misalignments. (B) Our Dream2Act pipeline leverages Seedance 2.0 to synthesize morphology-consistent interactions directly in the robot's native space. (C) This visual prior translates into successful zero-shot physical execution for spatially-sensitive tasks like sitting on a sofa and kicking a ball.}
    \label{fig:teaser}
\end{figure*}
The ultimate goal of embodied artificial intelligence is to deploy humanoid robots capable of executing general-purpose interaction tasks in unstructured human environments. To achieve this, acquiring dexterous interaction skills from demonstrations has emerged as a promising paradigm~\cite{fu2024humanplus, he2024omnih2o, ze2025twist}. 
However, capturing high-quality humanoid demonstrations conventionally relies on either expensive hardware-intensive teleoperation~\cite{fu2024mobile} or precise whole-body pose estimation for human-to-robot mapping~\cite{peng2018deepmimic}. 
Both pipelines are constrained by high hardware costs, substantial human labor, and a lack of scalability. 

While leveraging large-scale, in-the-wild human video data offers a more scalable alternative for skill acquisition~\cite{shen2024world, ni2025generated}, it introduces a fundamental challenge: how to bridge the physical mismatch between humans and robots.

To exploit human videos, mainstream approaches rely on human-to-robot motion retargeting~\cite{luo2023perpetual, araujo2025retargetingmattersgeneralmotion}. These paradigms typically utilize human-centric kinematic models, such as SMPL~\cite{loper2015smpl}, to extract poses from human videos and map them onto the robot-specific embodiment.

This human-centric pipeline is hindered by the morphology gap: humans and humanoids differ in skeletal proportions, degrees of freedom, and joint limits. Forcing human kinematics onto a robotic skeleton inevitably introduces severe spatial misalignments, a problem compounded by accumulated errors during locomotion. Even in fundamental whole-body mobile interactions, these scale mismatches and accumulated locomotion drifts severely warp the absolute spatial relationship between the agent and the target, consistently compromising task success.

In this work, we advocate a shift from human-centric retargeting to a strictly robot-centric alternative. We introduce Dream2Act, a novel framework that achieves zero-shot humanoid interaction through generative video synthesis. Unlike kinematic retargeting, which struggles to natively incorporate environmental geometry, recent generative video models like Seedance 2.0~\cite{seedance2_0, seedance2025seedance} have demonstrated the ability to implicitly capture physical dynamics and spatial-temporal consistency directly in the visual domain. Such models have been shown to encode scene affordances and inter-object relationships through large-scale visual pretraining~\cite{brooks2024video, du2023learning, bruce2024geniegenerativeinteractiveenvironments}. 
Given only a single third-person image of the humanoid and an interaction object, our framework leverages powerful image-to-video diffusion priors to hallucinate realistic and temporally coherent videos of the physical robot executing the desired task. 

Subsequently, we employ a high-fidelity pose estimation and reconstruction pipeline to recover robot-native kinematic trajectories directly from these synthesized sequences. Finally, these trajectories are executed through a general-purpose whole-body controller SONIC~\cite{luo2025sonic}. By entirely bypassing the the intermediate human embodiment representation, Dream2Act avoids retargeting-induced errors and sidesteps the morphology gap.

Our framework enables context-aware, task-level zero-shot interaction. By conditioning the generative process on a single real-world observation and a semantic task prompt, Dream2Act achieves the ability to execute novel tasks in a training-free manner, bypassing the need for any prior exposure to specific task dynamics or environment-specific data collection. We comprehensively validate Dream2Act across four representative categories of whole-body mobile interaction tasks deployed on the Unitree G1 humanoid. Extensive physical experiments demonstrate that while conventional optimization-based retargeting baselines suffer catastrophic failure in establishing correct physical contacts, our approach reliably achieves functional spatial alignment and successful physical engagement. 

Our work suggests that the large-scale generative video model ~\cite{seedance2_0} has internalized sufficient world physics to serve as a high-level zero-shot motion planner, paving the way for a new class of \textbf{hallucination-to-execution} paradigms in robotics. In summary, our main contributions are three-fold:
\begin{itemize}
    \item We propose Dream2Act, a pioneering robot-centric framework that enables task-level zero-shot humanoid interaction by leveraging generative video synthesis. This introduces a novel paradigm for whole-body robot motion acquisition, offering a scalable alternative to costly manual teleoperation.
    \item We introduce a robust kinematic recovery pipeline that effectively bridges the pixel-to-physics gap. By cascading native 2D pose estimation, a domain-specific 3D lifting network, and URDF-constrained inverse kinematics, our approach successfully translates generative visual priors into physically executable joint trajectories.
    \item We demonstrate the efficacy of Dream2Act on a physical Unitree G1 robot across four diverse categories of whole-body mobile interaction tasks. Our results establish a scalable, data-driven path for zero-shot embodied intelligence, reliably outperforming conventional optimization-based retargeting baselines.
\end{itemize}
\section{Related Work}
\label{sec:related}

\subsection{Human-Centric Motion Acquisition and Retargeting}
Imitating human demonstrations is a cornerstone for acquiring humanoid robotic skills. A prevalent paradigm relies on a multi-stage pipeline: first, human kinematics are extracted from monocular videos or motion capture data using human-centric parametric models such as SMPL~\cite{loper2015smpl} or GVHMR~\cite{shen2024world}. Similarly, in the Text-to-Motion domain, leading generative models, such as MDM~\cite{tevet2022human}, HY-Motion~\cite{wen2025hy}, Dart~\cite{gu2025dartdenoisingautoregressivetransformer}, and ViMogen~\cite{lin2025quest} from the T2M benchmark~\cite{yang2026t2mbench}, synthesize high-fidelity human kinematics directly from semantic prompts. 

Irrespective of the source, these human-centric poses must be mapped to robotic embodiments through optimization-based solvers like GMT~\cite{chen2025gmt} or GMR~\cite{araujo2025retargetingmattersgeneralmotion} to bridge inherent skeletal discrepancies. These retargeted trajectories serve as foundational reference motions for training RL policies or whole-body trackers, a methodology pioneered by DeepMimic~\cite{peng2018deepmimic}  and subsequently advanced by dexterous systems such as PHC~\cite{luo2023perpetual}, MimicKit~\cite{peng2025mimickit} OmniH2O~\cite{he2024omnih2o}, and HumanPlus~\cite{fu2024humanplus}. 

While effective for general locomotion, these methodologies are fundamentally bottlenecked by the morphology gap. Beyond simple discrepancies in skeletal proportions, these retargeting processes are often designed to optimize for kinematic resemblance in isolation from the physical scene. This leads to severe spatial misalignments and lost contact precision in object-centric tasks.

\subsection{Robot-Centric Generative Planning}
Recent breakthroughs in generative AI, particularly advanced large-scale video generative models like Wan 2.1~\cite{wan2025wan}, Sora~\cite{brooks2024video}, Stable Video Diffusion~\cite{blattmann2023stable}, and the state-of-the-art Seedance 2.0~\cite{seedance2_0}, have demonstrated unprecedented capabilities in modeling complex physical dynamics.  In the robotics community, extracting actionable priors from such models has gained immense traction. For tabletop robotic arms, extracting visual priors (e.g., R3m~\cite{nair2022r3m}) or regressing actions directly from human videos~\cite{shaw2024learning, xie2025human2robot} has proven highly effective.

For humanoids, recent works explore video world models like DreamGen~\cite{jang2025dreamgen} to unlock generalization, or frameworks like GenMimic~\cite{ni2025generated} that leverage generative models to synthesize human videos, which are then retargeted into robot trajectories. However, all these approaches still struggle with the aforementioned retargeting trap due to their reliance on human intermediaries. Dream2Act pioneers a strictly robot-centric approach: by prompting Seedance 2.0 to hallucinate the physical robot itself, we obtain a native temporal prior that bypasses retargeting entirely. This allows our framework to serve as an explicit zero-shot motion planner that is inherently grounded in the environment's geometry and the robot's native morphology.

\subsection{Robotic Perception and General-Purpose Tracking}
\label{sec:related_foundations}

Translating generative visual hallucinations into physical robot actions requires a robust integration of robotic pose perception and whole-body tracking~\cite{peng2018deepmimic}. While human pose models like HRNet~\cite{sun2019deep} and ViTPose~\cite{xu2022vitpose} are mature on datasets like COCO~\cite{lin2014microsoft}, they struggle with robotic embodiments due to the morphology gap and domain-specific data scarcity~\cite{xu2022vitpose}. Dream2Act overcomes this by leveraging simulation environments like IsaacLab~\cite{mittal2025isaac} and motion priors such as AMASS~\cite{mahmood2019amass} to synthesize a native dataset, enabling precise pose perception grounded in the robot's specific kinematic skeleton.

On the execution side, the field has evolved from task-specific imitation toward general-purpose whole-body tracking. Traditional frameworks often relied on analytical solvers like Pinocchio~\cite{carpentier2019pinocchio} or specialized systems like TWIST~\cite{ze2025twist}. While agile trackers such as BeyondMimic~\cite{liao2025beyondmimic} have improved responsiveness, they remain limited to specific motion distributions. Recently, the development of general tracking foundations—such as BFM-Zero~\cite{li2025bfm} and Sonic~\cite{luo2025sonic}—has enabled the tracking of diverse, complex motions across unstructured environments. Dream2Act synergizes with these foundations by providing morphology-consistent and kinematically plausible visual priors, achieving stable zero-shot physical execution without requiring additional task-specific training.

\section{Method}
\label{sec:method}
\begin{figure*}[t]
    \centering
    \includegraphics[width=\textwidth]{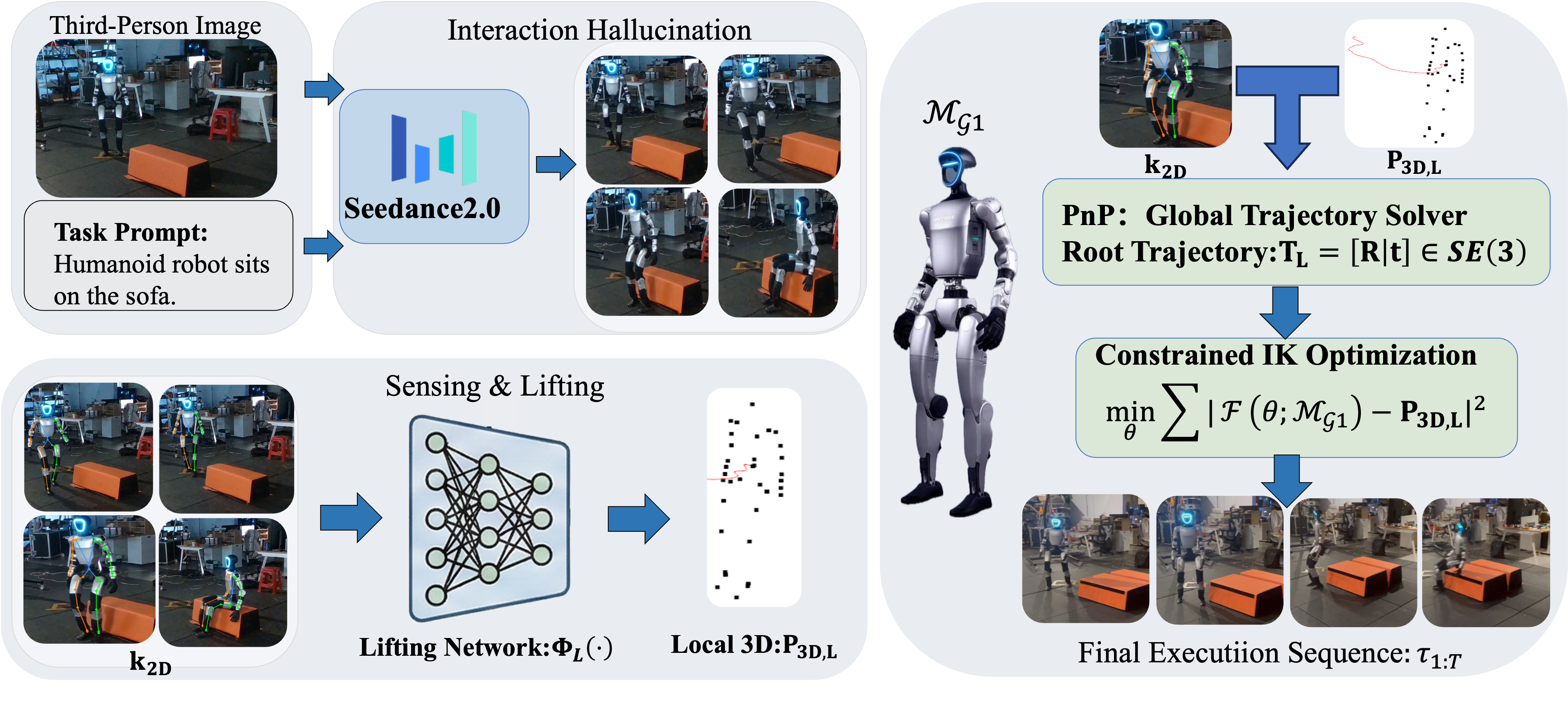}
    \caption{\textbf{Dream2Act System Architecture.} The framework comprises three core modules: (1) Interaction Hallucination via Seedance to generate morphology-consistent visual priors; (2) Native Pose Estimation and 2D-to-3D Lifting to extract precise 3D keypoints; and (3) Morphology-Aware Kinematic Recovery employing Constrained IK and PnP scale recovery to trajectories for physical execution.}
    \label{fig:pipeline}
\end{figure*}
The core objective of Dream2Act is to synthesize physically viable interaction trajectories directly from initial visual observations, bypassing explicit and error-prone human-centric retargeting paradigms. Operating through a strictly robot-centric pipeline, our framework comprises three primary stages: (1) Generative interaction hallucination via Seedance 2.0; (2) High-fidelity native pose estimation and 2D-to-3D lifting; and (3) Morphology-aware kinematic recovery and physical execution.

\subsection{Interaction Hallucination via Seedance 2.0}
\label{sec:video_generation}

To avoid the prohibitive costs associated with collecting complex interaction data in the physical world, we formulate motion planning as a video generation process. We employ the Seedance 2.0~\cite{seedance2_0} generative world model as our core predictive synthesis engine. Building upon the foundational architecture of its predecessor~\cite{seedance2025seedance}, Seedance 2.0 demonstrates unprecedented capabilities in modeling complex physical interactions and multi-subject dynamics. 

We select this model over alternatives due to its state-of-the-art performance in two critical dimensions required for robotic perception: (1) Physical Plausibility: It exhibits superior adherence to real-world physics (e.g., rigid body dynamics and contact constraints) during complex interactions; and (2) Morphological Consistency: It maintains exceptional structural rigidity of the subjects without introducing distortions or artifacts, even under significant viewpoint shifts or prolonged task execution.

Before executing any novel interaction task, our framework only requires a single initial frame captured by a third-person camera in the real environment, containing the physical Unitree G1 robot and the target object. This frame serves as the exact spatial conditioning. By incorporating text instructions or semantic task prompts, Seedance 2.0 synthesizes a physically intuitive and temporally coherent video that shows the robot successfully completing the intended task. Crucially, the generated video effectively maintains a fixed third-person camera perspective and preserves the physical dimensions of the G1 with high morphological fidelity. This high-quality, morphology-consistent video stream thereby serves as a reliable visual prior for the subsequent native kinematic recovery.
\subsection{Domain-Specific Robotic Dataset Construction}
\label{sec:dataset}

A fundamental bottleneck in deploying vision-based pose estimators on humanoid robots is the severe scarcity of domain-specific annotated datasets. To overcome this, we construct a large-scale, high-fidelity dataset for the G1 humanoid, integrating zero-cost simulation rendering with human-in-the-loop (HITL) real-world annotations.

\noindent\textbf{Simulation-based High-Fidelity Rendering.} To construct a robust motion prior, we leverage the AMASS dataset~\cite{mahmood2019amass}. We first retarget its extensive human motion sequences to the G1 29-DoF kinematic model, yielding a massive initial trajectory collection of 5,337,514 frames. Since highly correlated adjacent frames offer marginal information gain and degrade the learning efficiency of subsequent networks, we perform a rigorous temporal downsampling. This step aggressively distills the dataset into 5,338 diverse and representative keyframes for high-fidelity rendering.
With the robot anchored at the origin, we construct a hemispherical, omnidirectional multi-camera system in Isaac Lab. This array comprises nine high-resolution virtual cameras: four orthogonal side views, four diagonal views with a 45$^\circ$ downward pitch, and one strict top-down view.
To bridge the coordinate system gap between the physics simulation (USD convention) and standard computer vision pipelines, we perform rigorous orthogonal frame transformations. Using the derived camera extrinsics and intrinsics, we efficiently project the 3D world joints into local camera space and subsequently into precise 2D pixel coordinates $[u, v]^T$. Beyond skeletal joints, we synchronously render depth maps and instance segmentation masks to automatically extract tight bounding boxes and center points for the robot. This pipeline furnishes a rich, multi-modal ground-truth suite for the subsequent visual generation and lifting modules.

\noindent\textbf{In-the-Wild Data and Human-in-the-Loop 2D Annotation.} To bridge the Sim-to-Real gap for visual perception, we collect in-the-wild videos of the physical G1 robot and temporally downsample them by a factor of $10$ to reduce redundancy. We then partition these processed frames into a training set and an evaluation set. For the training set, we adopt a Human-in-the-Loop (HITL) annotation pipeline: a purely simulation-trained model first pre-annotates the 2D pixel coordinates, followed by expert manual correction of misaligned joints. Conversely, the evaluation set undergoes multi-round, fine-grained expert 2D annotation from scratch to establish a high-confidence benchmark. Crucially, this real-world dataset is exclusively utilized to fine-tune the 2D pose estimation network (detailed in Sec. \ref{sec:pose_estimation}) against real-world illumination and textures, whereas our 2D-to-3D lifting network relies on the exact 3D geometric ground truth provided by the simulation environment.

\subsection{Native Pose Estimation and 2D-to-3D Lifting}
\label{sec:pose_estimation}

Equipped with our hybrid dataset, we employ ViTPose~\cite{xu2022vitpose} as the backbone architecture for Parameter-Efficient Fine-Tuning (PEFT). We freeze the first nine layers of the pre-trained ViT-Base model and fine-tune only the final three layers alongside the top-down Gaussian heatmap head. Given an input image $I$, this estimator accurately predicts the 2D keypoints $P_{2D}$ for the $N=30$ native robotic joints.

To transition from the 2D pixel space to the 3D physical space, the extracted 2D keypoints are first normalized (zero-centered and scaled to $[-1, 1]$) to isolate global scale variations. Subsequently, these normalized 2D coordinates are fed into a specifically trained 2D-to-3D Lifting Network. This network explicitly models the native 3D joint distribution of the G1 robot, directly regressing the 3D relative joint coordinates $P_{3D}$ in the camera coordinate system, thereby reconstructing the native-scale 3D articulation required for subsequent kinematic recovery. 

Formally, the entire visual perception and lifting pipeline can be summarized as:
\begin{equation}
    P_{2D} = f_{\text{ViT}}(I), \quad P_{2D} \in \mathbb{R}^{N \times 2}
\end{equation}
\begin{equation}
    P_{3D} = \Phi_{\text{lift}}\big( \mathcal{N}(P_{2D}) \big), \quad P_{3D} \in \mathbb{R}^{N \times 3}
\end{equation}
where $f_{\text{ViT}}(\cdot)$ denotes the fine-tuned 2D pose estimator, $\mathcal{N}(\cdot)$ represents the spatial normalization function, and $\Phi_{\text{lift}}(\cdot)$ is the 2D-to-3D lifting network mapping the normalized 2D coordinates to the native-scale 3D domain.

\subsection{Morphology-Aware Kinematic Recovery and Physical Execution}
\label{sec:kinematic_recovery}

To recover the authentic motion trajectory of the G1 robot from the inferred 3D visual keypoints $P_{3D}$, we design a robust, decoupled pipeline that explicitly separates low-level internal joint kinematics from global root dynamics, culminating in authentic physical deployment.

\noindent\textbf{Stage 1: Internal Joint Kinematics via Constrained IK.} We first compute the relative positions $\hat{P}_{3D}$ by centering all 3D points relative to the pelvis. The recovery of joint angles is formulated as a constrained Inverse Kinematics (IK) optimization problem:
\begin{equation}
    q^{*} = \arg\min_{q} \sum_{i=1}^{N} \big\| \mathrm{FK}_{i}(q) - \hat{P}_{3D}^{(i)} \big\|^{2}
\end{equation}
subject to joint limits $q_{min} \le q \le q_{max}$. By fixing the root node at the origin and optimizing solely on position constraints, the IK solver generates a structurally consistent internal joint configuration $q^{*}$ that strictly adheres to the rigid URDF model of the G1.

\noindent\textbf{Stage 2: Scale Recovery and Root Pose Optimization.} We apply Forward Kinematics (FK) to the internal state $q^{*}$ (with the root fixed as an identity transformation) to obtain precise local 3D points $P_{local} \in \mathbb{R}^{N \times 3}$. Subsequently, we align these points with the un-normalized original 2D detections $P_{2D}$ using the Perspective-n-Point (PnP) algorithm:
\begin{equation}
    \min_{R, t} \sum_{i=1}^{N} \big\| P_{2D}^{(i)} - \pi \big( K(R \cdot P_{local}^{(i)} + t) \big) \big\|^{2}
\end{equation}
Since $P_{local}$ inherits the exact physical dimensions from the URDF model, the optimized translation vector $t$ and rotation matrix $R$ recover the true metric scale and global pose. Finally, an Exponential Moving Average (EMA) filter is applied to temporally smooth the trajectories of $R$ and $t$. Subsequently, a rigorous coordinate transformation is performed to project these global trajectories back into the robot's egocentric base frame, yielding the native motion sequence for downstream physical execution.

\noindent\textbf{Stage 3: Unified Formatting and Sonic Tracker Execution.} Upon completing the kinematic recovery, we encapsulate the optimized global translation $t$, global rotation $R$ (converted to quaternions), and internal joint angles $q^{*}$ into the standardized General Motion Retargeting (GMR) format, yielding a unified high-dimensional state vector \texttt{(root\_pos, root\_rot, qpos)}. This standardized motion representation facilitates immediate zero-shot real-world deployment. We directly feed the generated GMR trajectory into the open-source Sonic general-purpose motion tracker. The Sonic tracker manages the low-level Whole-Body Control (WBC) and balance maintenance, driving the physical G1 robot to accurately track the extracted motion and end-to-end execute complex interaction tasks in real-world settings.

\section{Experiments}
\label{sec:experiments}

Our experiments are systematically designed to answer the following research questions: (1) Can Dream2Act enable zero-shot physical interactions compared to human-centric retargeting? (2) Does robot-centric video synthesis offer a more physically plausible paradigm for free-space motion generation than conventional text-to-motion models? (3) how our specific system designs, namely the hybrid dataset, contribute to robust perception.?

\subsection{Experimental Setup and Datasets}
\label{sec:exp_setup}
\noindent\textbf{Datasets Construction.} Training accurate native pose estimators requires domain-specific data. We construct a hybrid dataset comprising two distinct domains: (1) Simulation Dataset: We leverage AMASS motion priors, retargeting them to the G1 to yield over 5.3 million frames, which are then rigorously downsampled to 5,338 diverse keyframes. Through a 9-camera hemispherical setup in Isaac Lab, we render high-fidelity images accompanied by exact 3D geometric ground truth ($P_{world}$), bounding boxes, and masks. (2) In-the-Wild Real Dataset: To address the Sim-to-Real visual domain shift, we collect real-world videos of the G1, temporally downsampled to a factor of 10. These frames are partitioned into a 90\% training set (refined via Human-In-The-Loop expert correction) and a 10\% evaluation set (341 frames, annotated from scratch via multi-round expert labeling) to serve as a high-confidence benchmark.

\begin{figure}[t]
    \centering
    \includegraphics[width=\linewidth]{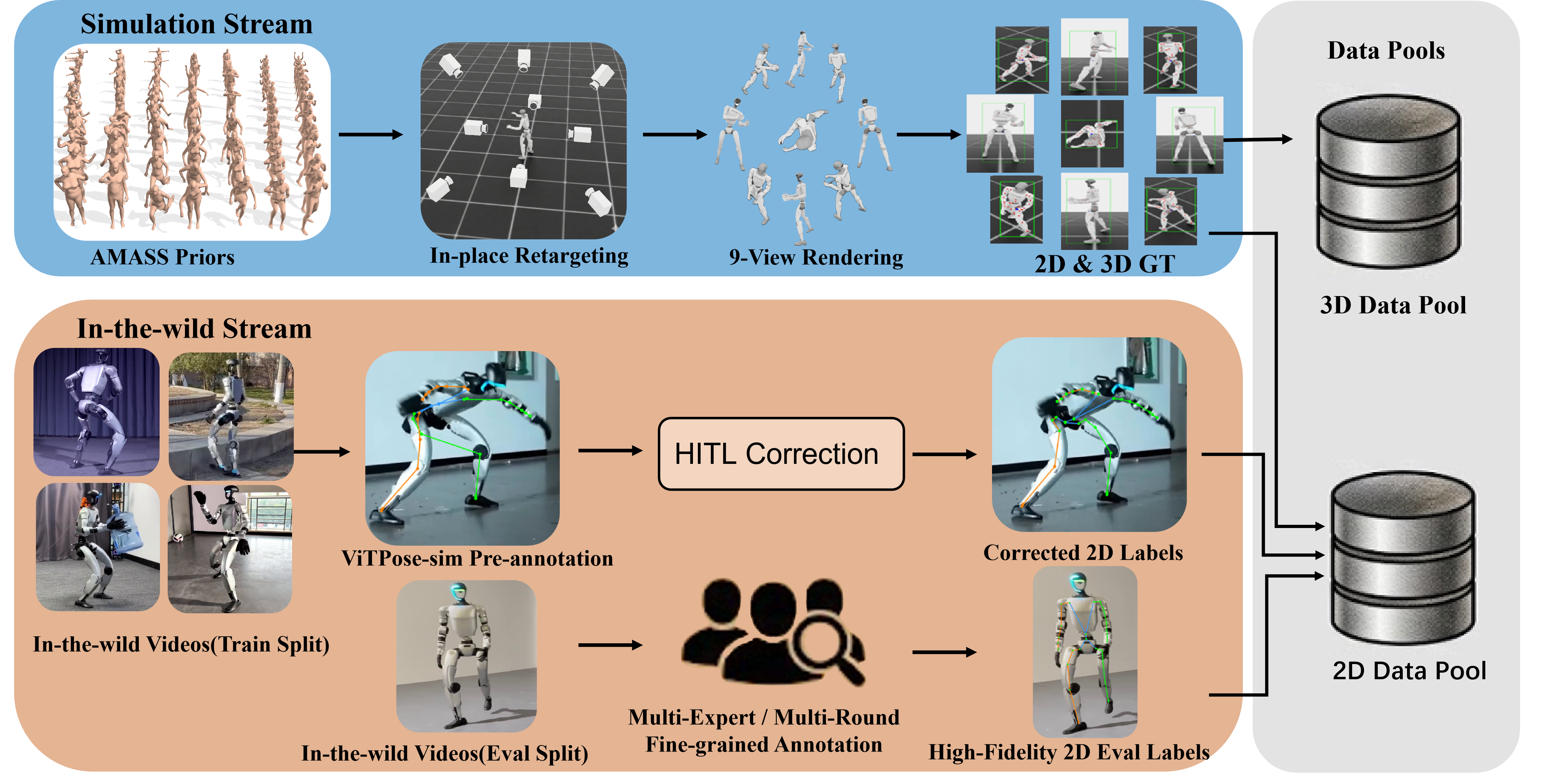}
    \caption{\textbf{G1 Domain-Specific Hybrid Dataset Construction Pipeline.} Our approach synergizes cost-free, high-fidelity 3D/2D geometry rendered from a 9-camera simulation environment (Isaac Lab) with Human-in-the-Loop (HITL) annotations to correct visual domain shifts, enabling robust native pose estimation.}
    \label{fig:dataset}
\end{figure}

\noindent\textbf{Hardware and Execution.} All physical deployments are conducted on the Unitree G1 humanoid (29 DoF). The extracted unified trajectories are seamlessly executed via the open-source Sonic general-purpose whole-body controller ~\cite{luo2025sonic} running at 50 Hz. 

\subsection{Real-World Zero-Shot Interaction}
\label{sec:exp_interaction}
\begin{figure*}[t]
    \centering
    \includegraphics[width=\textwidth]{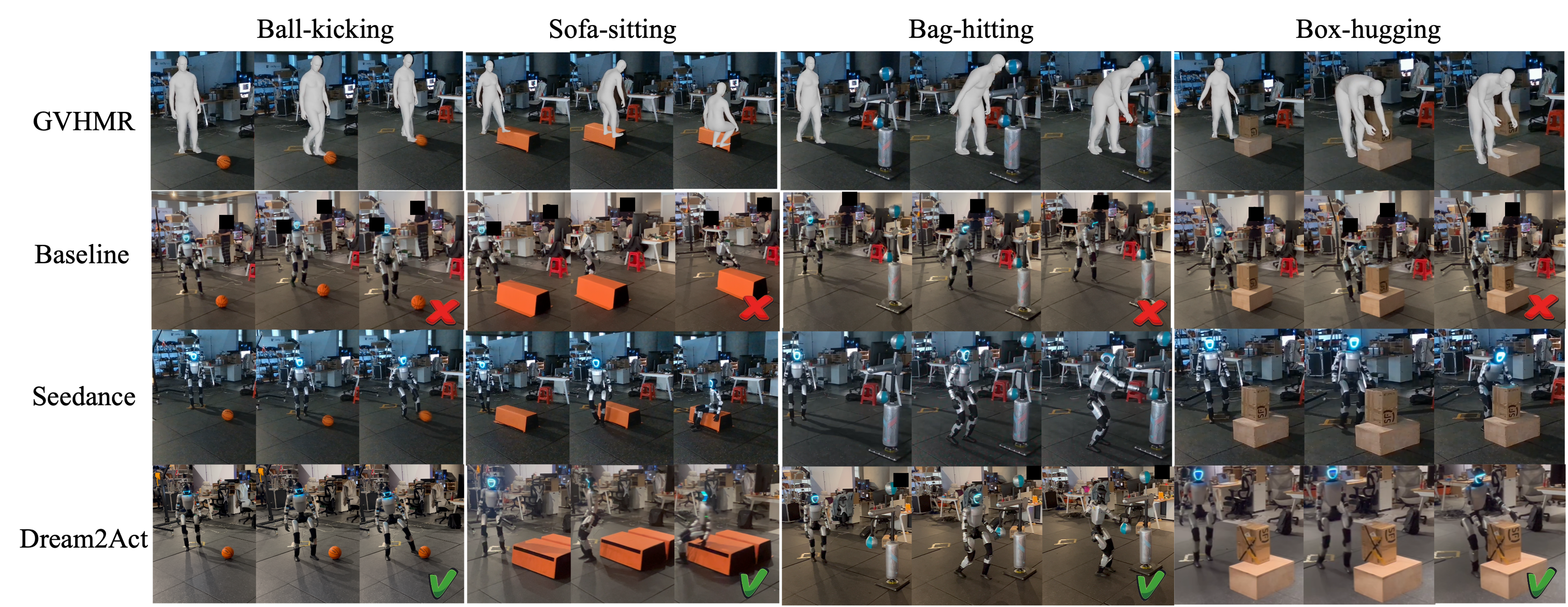}
    \caption{\textbf{Qualitative Comparison on Diverse Spatially-Sensitive Tasks.} We evaluate zero-shot performance across Kicking, Hugging, Punching, and Sitting. (Top) Traditional human-centric retargeting suffers from spatial misalignments and kinematic violations due to the morphology gap. (Bottom) Dream2Act's robot-centric hallucination ensures precise contacts and dynamic stability.}
    \label{fig:interaction_comparison_main}
\end{figure*}

To evaluate morphology-aware interaction, we define four diverse, spatially-sensitive tasks: (1) Ball-Kicking; (2) Box-hugging; (3) Bag-hitting; and (4) Sofa-sitting.

\noindent\textbf{Baselines.} To strictly isolate the impact of the morphology gap, we compare our approach against a representative optimization-based retargeting baseline: Baseline (Real Human Video): This pipeline extracts human kinematics from real-world demonstration videos using GVHMR~\cite{shen2024world}, which are subsequently mapped to the G1 via GMR~\cite{araujo2025retargetingmattersgeneralmotion} and executed by the Sonic~\cite{luo2025sonic} controller.

In contrast, Dream2Act (ours) operates through a strictly robot-centric paradigm. By prompting Seedance 2.0 to hallucinate the G1 robot directly, our framework bypasses the need for human-to-robot mapping. The synthesized native motion is processed through our domain-specific 2D-to-3D lifting and URDF-constrained IK optimization (Stage 2 \& 3) before being tracked by the Sonic whole-body controller. 

\noindent\textbf{Metrics and Results.} We report the Success Rate and the Average Spatial Alignment Error (calculated as the Mean Absolute Error of the end-effector distance to the intended contact point) across 20 trials per task. To evaluate robustness over distance, we also report the initial Task Length (the physical distance from the robot's starting position to the interaction target). As shown in Table~\ref{tab:interaction_main} and Table~\ref{tab:interaction_kicking}, the human-centric baseline fundamentally fails on spatially-sensitive tasks. Furthermore, we observe that traditional retargeting errors are severely compounded under locomotion. For instance, in the 'Kick Ball (d)' task requiring 2.18m of mobility, the human-centric baseline accumulates an overwhelming 0.79m of spatial misalignment. In contrast, Dream2Act effectively bounds the execution error (averaging 0.14m across kicking tasks) regardless of the task length. By operating natively within the G1's embodiment, Dream2Act preserves precise spatial relationships and absolute metric scales, achieving significantly higher success rates and robust task completion.

\begin{table}[t]
\centering
\caption{Quantitative comparison on four zero-shot interaction tasks. Task Length (in italics) denotes the initial distance to the target. Dream2Act significantly outperforms human-centric baselines, effectively bounding spatial alignment errors.}
\label{tab:interaction_main}
\resizebox{\linewidth}{!}{
\begin{tabular}{l|cc|cc|cc|cc}
\toprule
\multirow{2}{*}{\textbf{Method}} & \multicolumn{2}{c|}{\textbf{Ball-kicking} (\textit{1.64m Avg})} & \multicolumn{2}{c|}{\textbf{Box-hugging} (\textit{0.75m})} & \multicolumn{2}{c|}{\textbf{Bag-hitting} (\textit{1.60m})} & \multicolumn{2}{c}{\textbf{Sofa-sitting} (\textit{1.10m})} \\
 & Succ $\uparrow$ & Err $\downarrow$(m) & Succ $\uparrow$ & Err $\downarrow$(m) & Succ $\uparrow$ & Err $\downarrow$(m) & Succ $\uparrow$ & Err $\downarrow$(m) \\
\midrule
Baseline & 0\% & 0.75 & 0\% & 0.35 & 0\% & 0.70 & 0\% & 0.65 \\
\textbf{Dream2Act (Ours)} & \textbf{40\%} & \textbf{0.14} & \textbf{10\%} & \textbf{0.11} & \textbf{30\%} & \textbf{0.10} & \textbf{70\%} & \textbf{0.28} \\
\bottomrule
\end{tabular}
}
\end{table}

\begin{figure*}[t]
    \centering
    \includegraphics[width=\textwidth]{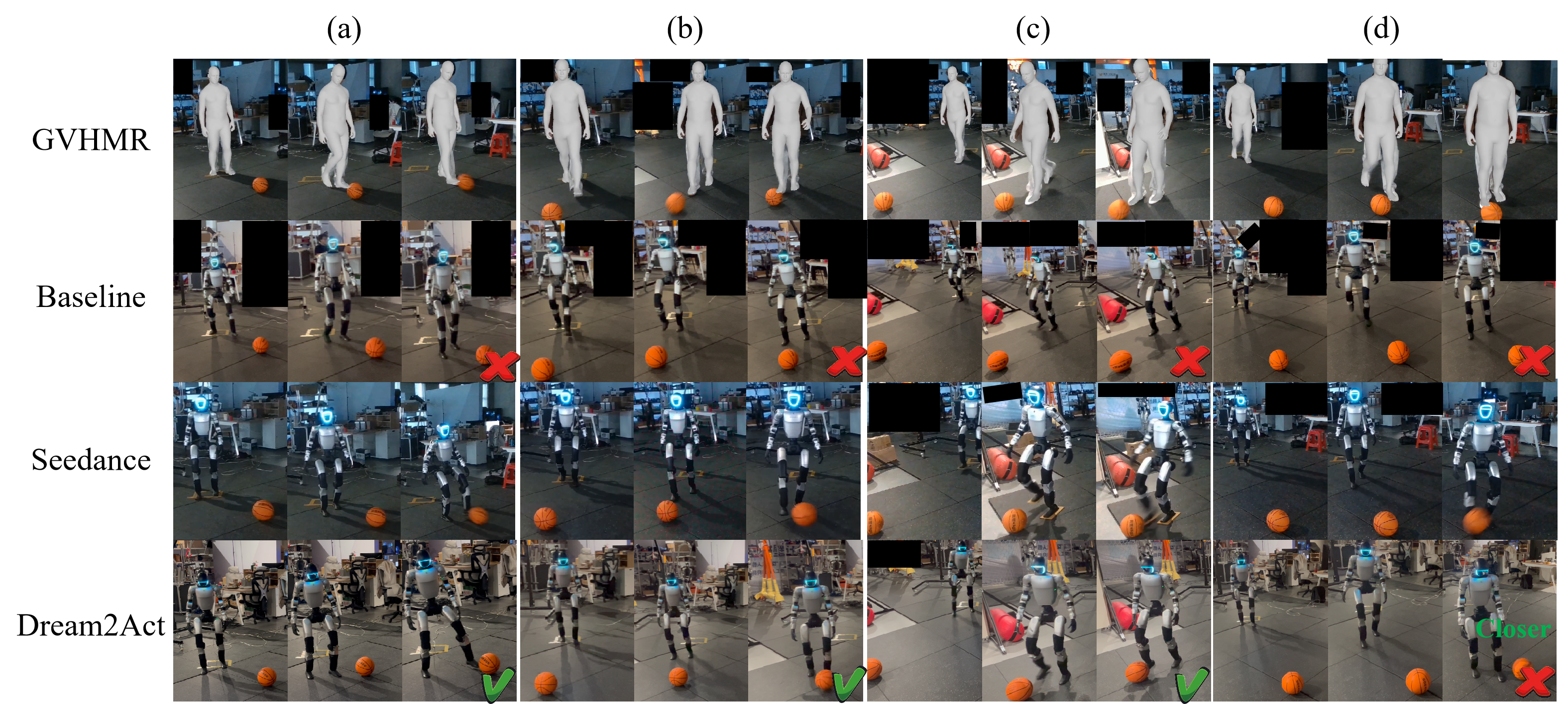}
    \caption{\textbf{Spatial Generalization Analysis on Multi-Location Ball Kicking.} To verify scale adaptability, we evaluate Dream2Act across four distinct ball positions. (Top) Human-centric retargeting fails to replicate spatial variations due to the morphology gap. (Bottom) Dream2Act successfully hallucinates and executes kicks for all targets, demonstrating robust spatial diversity.}
    \label{fig:interaction_comparison_kicking}
\end{figure*}

\begin{table}[t]
\centering
\caption{Quantitative comparison on multi-location ball-kicking. Task lengths vary from 0.90m to 2.18m. Dream2Act maintains precise spatial alignment across all targets, avoiding the error accumulation seen in baselines.}
\label{tab:interaction_kicking}
\resizebox{\linewidth}{!}{
\begin{tabular}{l|cc|cc|cc|cc}
\toprule
\multirow{2}{*}{\textbf{Method}} & \multicolumn{2}{c|}{\textbf{Kick Ball (a)} (\textit{0.90m})} & \multicolumn{2}{c|}{\textbf{Kick Ball (b)} (\textit{2.05m})} & \multicolumn{2}{c|}{\textbf{Kick Ball (c)} (\textit{1.45m})} & \multicolumn{2}{c}{\textbf{Kick Ball (d)} (\textit{2.18m})} \\
 & Succ $\uparrow$ & Err $\downarrow$(m) & Succ $\uparrow$ & Err $\downarrow$(m) & Succ $\uparrow$ & Err $\downarrow$(m) & Succ $\uparrow$ & Err $\downarrow$(m) \\
\midrule
Baseline & 0\% & 0.53 & 0\% & 0.82 & 0\% & 0.84 & 0\% & 0.79 \\
\textbf{Dream2Act (Ours)} & \textbf{60\%} & \textbf{0.19} & \textbf{20\%} & \textbf{0.09} & \textbf{60\%} & \textbf{0.09} & \textbf{20\%} & \textbf{0.19} \\
\bottomrule
\end{tabular}
}
\end{table}

\subsection{Text-Driven Free-Space Motion Generation}
\label{sec:exp_t2m}
\begin{figure}[t]
    \centering
    \includegraphics[width=\linewidth]{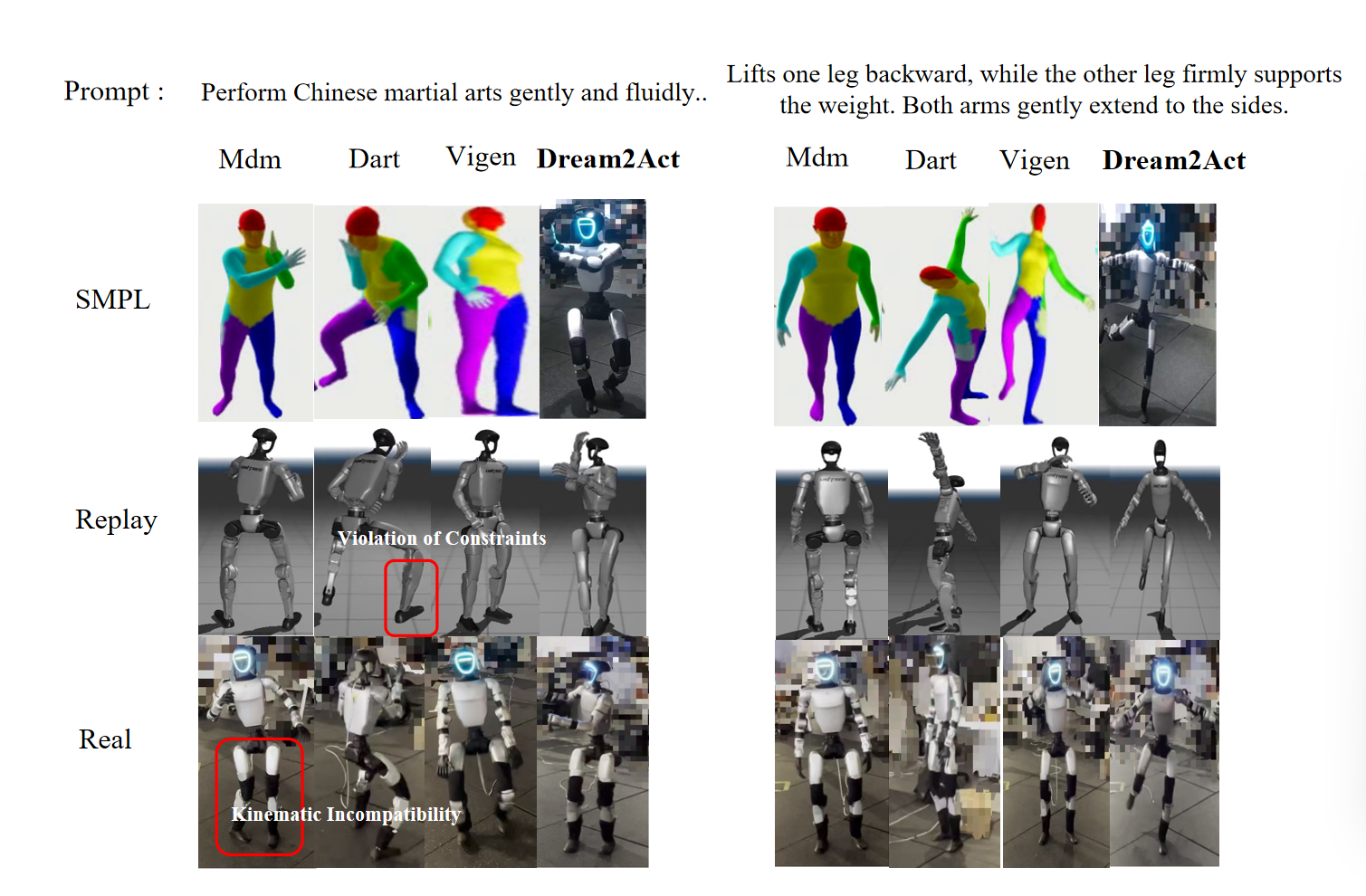}
    \caption{\textbf{Qualitative Comparison of Text-to-Motion Pipelines in Free Space.} Human-centric T2M models amplify inherent artifacts during retargeting, causing dynamic constraint violations and hardware failure. Dream2Act strictly enforces physical consistency, achieving stable zero-shot deployment.}
    \label{fig:t2m_comparison}
\end{figure}

Beyond object-centric interaction, we qualitatively evaluate the efficacy of our robot-centric synthesis against conventional human-centric Text-to-Motion (T2M) pipelines. Given identical semantic prompts, baselines utilize state-of-the-art T2M models, such as ViMogen~\cite{lin2025quest}, MDM~\cite{tevet2022human}, and Dart~\cite{gu2025dartdenoisingautoregressivetransformer}, to generate human kinematics, which are subsequently retargeted via GMR for robot deployment. 

\begin{figure*}[t]
    \centering
    \includegraphics[width=0.8\textwidth]{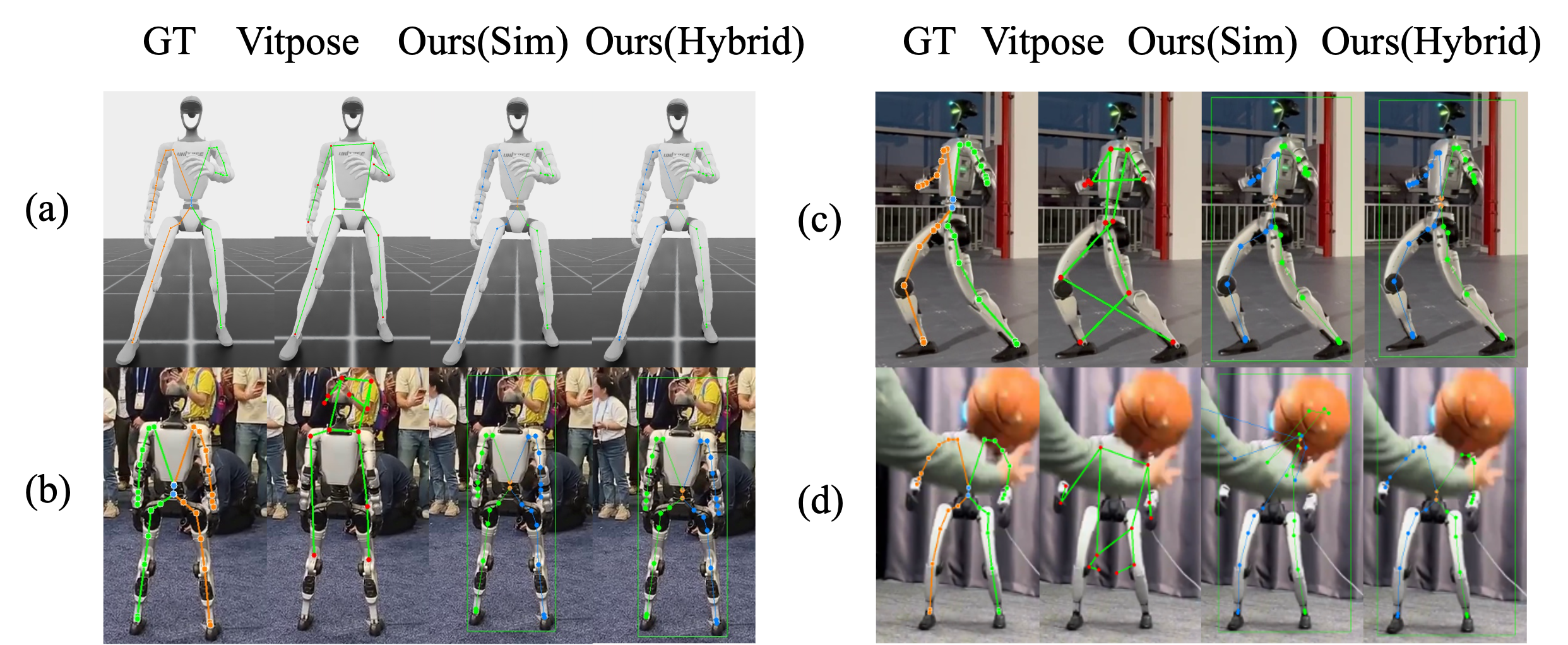}
    \caption{\textbf{Qualitative Ablation on Native G1 Pose Estimation.} Given a challenging real-world image: (a) Zero-shot ViTPose (COCO-only) completely misidentifies robot features; (b) A Sim-Only variant suffers from local drift due to domain shift; (c) Our Dream2Act native estimator, fine-tuned on the hybrid dataset, achieves pixel-perfect alignment.}
    \label{fig:pose_ablation}
\end{figure*}

\noindent\textbf{Qualitative Results and Analysis.} Our qualitative assessment, illustrated in Fig.~\ref{fig:t2m_comparison}, reveals that traditional human-centric pipelines suffer from severe compounding errors that manifest in three critical failure modes. First, kinematic incompatibility frequently arises because generated human poses often involve joint configurations or degrees of freedom that do not exist on the G1 robot, causing the mapping process to fail and resulting in frozen motor execution. Furthermore, these baselines consistently violate physical support constraints by hallucinating aerial trajectories where the agent appears to float or translate in mid-air without appropriate ground contact, a phenomenon that is physically unrealizable for a ground-based humanoid. Finally, minor artifacts inherent in human motion generation, such as foot sliding or jitter, are catastrophically amplified during retargeting, leading to violations of the G1's dynamic balance constraints and causing the physical robot to fall. In contrast, Seedance 2.0 generates motions that are implicitly grounded in the G1's native physical embodiment. By synthesizing morphology-consistent trajectories from the start, Dream2Act ensures that the resulting motion flawlessly maintains balance and physical stability while strictly adhering to the semantic intent of the text prompt.

\subsection{Module Evaluation I: Native Pose Perception}
\label{sec:exp_perception}
Robust 2D perception is the prerequisite for kinematic recovery. We evaluate our 2D detection module on both the Simulation Test Set and the In-the-Wild Internet Test Set.
\noindent\textbf{Baselines and Ablation.} We compare against zero-shot human pose estimators (ViTPose~\cite{xu2022vitpose} pre-trained on COCO). We ablate our training strategy by comparing a variant fine-tuned solely on simulation data (\textit{Ours-Sim Only}) against our model trained on the hybrid dataset (\textit{Ours-Hybrid}).

\noindent\textbf{Results.} As detailed in Table \ref{tab:perception}, off-the-shelf human detectors fail catastrophically on the G1 morphology. While the \textit{Sim Only} variant performs flawlessly on simulated images, its Average Precision (AP) and 2D PCK drop precipitously on the Internet test set due to lighting and texture domain shifts. Our \textit{Hybrid} model effectively closes this Sim-to-Real gap, achieving robust native joint localization across all environments. 

\begin{table}[h]
\centering
\caption{2D Pose Estimation Performance on 12 Common Joints. We compare zero-shot generalization of standard human estimators (COCO-17) vs. our domain-specific model on shared kinematic joints.}
\label{tab:perception}
\resizebox{0.9\linewidth}{!}{
\begin{tabular}{l|cc|cc}
\toprule
\multirow{2}{*}{\textbf{Method}} & \multicolumn{2}{c|}{\textbf{Simulation Test Set (2402 frames)}} & \multicolumn{2}{c}{\textbf{Internet Test Set (341 frames)}} \\
 & \textbf{AP} $\uparrow$ & \textbf{2D PCK@0.05} $\uparrow$ & \textbf{AP} $\uparrow$ & \textbf{2D PCK@0.05} $\uparrow$ \\
\midrule
Zero-shot ViTPose & 0.234 & 64.99\% & 0.260 & 66.06\% \\
\midrule
Ours (Sim Only) & \textbf{0.940} & \textbf{99.76\%} & 0.497 & 91.61\% \\
\textbf{Ours (Hybrid)} & 0.785 & 95.78\% & \textbf{0.613} & \textbf{97.02\%} \\
\bottomrule
\end{tabular}
}
\end{table}

\subsection{Module Evaluation II: 2D-to-3D Lifting}
\label{sec:exp_lifting}
The precision of converting 2D pixel coordinates $\mathbf{k}_{2D}$ into 3D joint positions $\mathbf{P}_{3D,L}$ is fundamental to the fidelity of the final humanoid motion. To quantify the performance of our 2D-to-3D lifting network $\Phi_L(\cdot)$, we conduct a rigorous evaluation on the simulation test set, which provides absolute 3D ground truth for the G1 robot $\mathcal{M}_{G1}$. Using Mean Per Joint Position Error (MPJPE) as the primary metric, our lifting network achieves a low average error of 29mm on the simulation benchmark. This high precision confirms the network's capability to accurately regress the robot's native-scale 3D articulation from monocular keypoints, effectively overcoming the depth ambiguity inherent in single-view perception by explicitly modeling G1-specific joint distributions. Such high-fidelity 3D reconstruction provides a geometrically consistent prior for the subsequent URDF-constrained IK solver, ensuring that the recovered trajectories $\tau_{1:T}$ not only visually align with the generative hallucinations but also strictly adhere to the robot's physical dimensions for stable real-world deployment.
\section{Conclusion}
\label{sec:conclusion}
In this paper, we presented Dream2Act, a pioneering robot-centric framework that enables true zero-shot humanoid interaction through generative video synthesis. By shifting from traditional human-centric retargeting to directly synthesizing interactions in the robot's native spatial coordinate system, our approach fundamentally eliminates the persistent morphological gap and spatial misalignments that have long bottlenecked dexterous skill acquisition. To seamlessly bridge the pixel-to-physics gap, we introduced a robust, domain-specific kinematic recovery pipeline—cascading native 2D pose estimation, 2D-to-3D lifting, and URDF-constrained inverse kinematics—to translate visual priors into physically viable joint trajectories. Extensive physical validations on the Unitree G1 humanoid across a diverse set of spatially-sensitive interactions (e.g., ball kicking, sofa sitting, and bag punching) demonstrate that Dream2Act reliably achieves correct physical contacts and significantly outperforms conventional optimization-based baselines in both spatial alignment and task success rates. While currently functioning as an offline zero-shot motion planner due to the inherent inference latency of diffusion models, Dream2Act bypasses prohibitive data collection and task-specific policy training, establishing a highly scalable, data-driven pathway for advancing embodied humanoid intelligence, with future work focusing on accelerating inference for real-time closed-loop control.

\clearpage
{
    \small
    \bibliographystyle{cvpr}
    \bibliography{ref}
}



\end{document}